\documentclass[10pt,twocolumn,letterpaper]{article}

\usepackage{3dv}
\usepackage{times}
\usepackage{epsfig}
\usepackage{graphicx}
\usepackage{amsmath}
\usepackage{amssymb}

\makeatletter
\@namedef{ver@everyshi.sty}{}
\makeatother
\usepackage{tikz}
\usepackage{booktabs}
\usepackage{adjustbox}
\usepackage{cite} 
\usepackage{makecell}
\usepackage{xcolor}
\usepackage{colortbl}
\usepackage{dashrule}
\usepackage{mathtools}
\usepackage{subfigure}
\usepackage{float}
\usepackage{enumitem}

\usepackage{hyperref}
\hypersetup{pagebackref=true,breaklinks=true,letterpaper=true,colorlinks,bookmarks=false}

\threedvfinalcopy 

\def\threedvPaperID{29} 
\def\httilde{\mbox{\tt\raisebox{-.5ex}{\symbol{126}}}}

\newcommand*{\defeq}{\stackrel{\text{def}}{=}}
\newcommand{\tabitem}{~~\llap{\textbullet}~~}

\DeclareMathOperator{\tr}{tr}
\makeatletter
\newcommand\footnoteref[1]{\protected@xdef\@thefnmark{\ref{#1}}\@footnotemark}
\makeatother

\newcommand\blfootnote[1]{%
  \begingroup
  \renewcommand\thefootnote{}\footnote{#1}%
  \addtocounter{footnote}{-1}%
  \endgroup}

\newcommand{\startcompact}[1]{\par\vspace{-0.75em}\begin{#1}%
\allowdisplaybreaks\ignorespaces}
\newcommand{\stopcompact}[1]{\end{#1}\ignorespaces}

\newenvironment{packed_item}{
\begin{itemize}[noitemsep,topsep=0pt]
   \setlength{\itemsep}{4pt}
   \setlength{\parskip}{0pt}
   \setlength{\parsep}{0pt}
 }{\end{itemize}}

\definecolor{sol_light_blue}{RGB}{38, 139, 210}
\definecolor{sol_blue}{RGB}{38, 139, 210}
\definecolor{nord_blue}{RGB}{38, 139, 210}
\definecolor{sol_green}{RGB}{163, 190, 140}
\definecolor{sol_red}{RGB}{220, 50, 47}
\definecolor{nord_red}{RGB}{250, 190, 192}
\definecolor{nord_green}{RGB}{163, 190, 140}

\definecolor{nordblack}{RGB}{46, 52, 64}
\definecolor{nordred}{RGB}{191, 97, 106}
\definecolor{nordgreen}{RGB}{163, 190, 140}
\definecolor{nordblue}{RGB}{94, 129, 172}
\definecolor{nordpurple}{RGB}{180, 142, 173}

\usetikzlibrary{patterns}
\DeclareRobustCommand{\legendsquare}[1]{%
  \tikz[baseline=(a.south)]{\node[#1, inner sep=.8ex, outer sep=0] (a) {};}%
}

\begin{document}

\title{Scene Flow from Point Clouds with or without Learning}

\author{Jhony Kaesemodel Pontes$^{1}$ \quad James Hays$^{1,2}$ \quad Simon Lucey$^{3}$ \\
\begin{tabular}[h]{cc}
	$^{1}$Argo AI \quad $^{2}$Georgia Institute of Technology \quad $^{3}$The University of Adelaide
\end{tabular}      
}


\twocolumn[{%
\renewcommand\twocolumn[1][]{#1}%
\maketitle
\vspace{-1cm}
\begin{figure}[H]
    \setlength{\hsize}{\textwidth}
    \includegraphics[width=\textwidth,keepaspectratio]{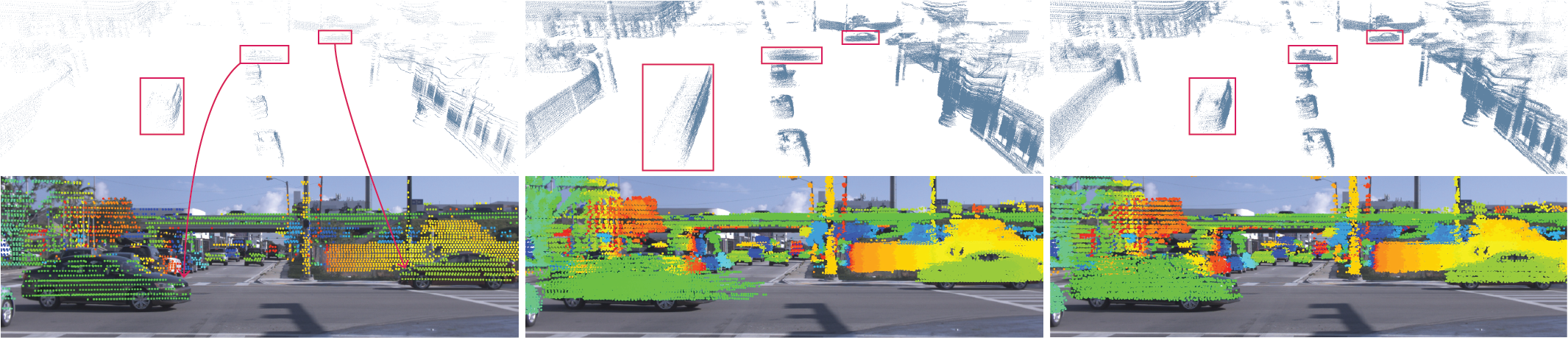}
    \caption{\textbf{Application of our scene flow approach for point cloud densification.} \textbf{Left:} Sparse point cloud and its projection onto an image. \textbf{Middle:} Densification using iterative closest point (ICP) which is a global rigid registration method. Five adjacent point clouds in each direction were used to densify the current frame. Note the ``ghosting'' effect due to the moving objects in the scene. \textbf{Right:} Densification using our method. Example from the Argoverse dataset. Colors repeat from green (close) to red (far) every fifty meters.}
    \label{fig:accumulation}
\end{figure}%
}]


\begin{abstract}
Scene flow is the three-dimensional (3D) motion field of a scene. It provides information about the spatial arrangement and rate of change of objects in dynamic environments. Current learning-based approaches seek to estimate the scene flow directly from point clouds and have achieved state-of-the-art performance. However, supervised learning methods are inherently domain specific and require a large amount of labeled data. Annotation of scene flow on real-world point clouds is expensive and challenging, and the lack of such datasets has recently sparked interest in self-supervised learning methods. How to accurately and robustly learn scene flow representations without labeled real-world data is still an open problem. Here we present a simple and interpretable objective function to recover the scene flow from point clouds. We use the graph Laplacian of a point cloud to regularize the scene flow to be ``as-rigid-as-possible''. Our proposed objective function can be used with or without learning---as a self-supervisory signal to learn scene flow representations, or as a non-learning-based method in which the scene flow is optimized during runtime. Our approach outperforms related works in many datasets. We also show the immediate applications of our proposed method for two applications: motion segmentation and point cloud densification.
\end{abstract}

\section{Introduction}
Supervised learning approaches rely on a large amount of labeled data which is not always readily available. Also, the learned models are domain specific and do not generalize well to other scenarios that are statistically different from the training data. For example, a model trained to predict scene flow on an unrealistic synthetic dataset will likely not perform well on a real self-driving situation. Current supervised methods that estimate scene flow from point clouds~\cite{liu19, wang20, gwwlw19} are trained on large-scale synthetic data, \eg FlyingThings3D~\cite{MIFDB16}, to be further fine-tuned on small real-world datasets such as the KITTI Scene Flow~\cite{mg15, mhg15}. Although there is an increasing availability of large-scale self-driving datasets with point cloud data from light detection and ranging sensors (lidar)~\cite{argoverse, nuscenes2019, waymo2020}, they do not provide scene flow labels. Annotation of scene flow on lidar data is challenging and expensive, given that a pair of point clouds needs translational vector labels describing the motion at every point in the scene. 

As a result, self-supervised methods have recently been proposed~\cite{wu19, mittal2019just} to learn scene flow representations without leveraging human supervision. Given two consecutive point clouds, the idea is to estimate how each point in the source point cloud moves towards its corresponding point on the next point cloud. The collection of all the individual motions, or translational vectors, is the scene flow. A global rigid transformation to represent dynamic scene flow is not sufficient---instead there is a transformation for every point. However, solving for individual transformations is an ill-posed problem~\cite{NICP}. Most of the prior work aims at improving scene flow estimation by using deep hierarchical networks~\cite{liu19, wang20, gwwlw19, wu19, mittal2019just}. They perform recursive sampling and grouping of neighboring points in different scales to regularize the scene flow to be similar in local regions.

In this work, we propose a simple and geometrically interpretable objective function to optimize the scene flow. We constrain the scene flow such that points that are close to each other in a local region move rigidly while the collection of all points move ``as-rigid-as-possible''~\cite{arap2007}. Such intuition is elegantly captured by the widely used graph Laplacian~\cite{s05, arap2007, bobenko2007, Keenan2015, keenan2019, eisenberger2019, Wang2018_eccv} which implicitly embeds the topological structure of a point cloud. Therefore, we propose the formation of an explicit graph on the source point cloud to capture its topology and context information about how points are locally connected. The graph Laplacian representation allows us to regularize and robustly approximate the scene flow from point clouds without annotations. 

Our proposed objective function can be used for self-supervised learning or for non-parametric optimization (for clarity, we exchange \textit{non-parametric} to \textit{non-learning} from now on). We evaluate our method on the synthetic FlyingThings3D~\cite{MIFDB16} dataset and on the real-world KITTI Scene Flow~\cite{mg15, mhg15}, Argoverse~\cite{argoverse}, and nuScenes~\cite{nuscenes2019} datasets. Since Argoverse and nuScenes do not provide scene flow annotations, we use the ego-vehicle poses and 3D object tracks to create pseudo labels. 

Our approach outperforms prior self-supervised works on many datasets \textit{with or without learning}. It not only provides a robust self-supervisory signal to learn scene flow representation but is also a robust non-learning objective for runtime optimization. Furthermore, we explore the applications of our method for two applications: motion segmentation and point cloud densification.

\textbf{Our main contributions are:}
\begin{packed_item}
    \item a simple and geometrically interpretable objective function to approximate scene flow from a pair of point clouds. An ``as-rigid-as-possible'' regularizer constrains the non-rigid motion flow based on the graph Laplacian of the source point cloud;
    \item our objective function can be used with or without learning. As a self-supervisory signal to learn scene flow representations, or as a non-learning method in which the scene flow is optimized during runtime;
    \item we show compelling results on synthetic, FlyingThings3D, and real-world datasets, KITTI Scene Flow, Argoverse, and nuScenes---with or without learning.
\end{packed_item}

\section{Related Work}
Here we review the most relevant scene flow works based on two broad categories: non-learning-based and learning-based scene flow methods. We focus on point-based scene flow, but for completeness we also include a review of image-based scene flow methods. Moreover, we briefly review related works that make use of the graph Laplacian.

\subsection{Non-learning-based scene flow}
\noindent\textbf{Image-based scene flow.}~
The development of non-learning scene flow methods is traced back to the classical work of Vedula~\etal~\cite{vbrck99}. They represented the scene flow as a dense vector field defined for every point on every surface in the scene. Then, a two-step approach was proposed to estimate the scene flow in a decoupled way: $(1)$ optical flow is computed for the multi-view image sequence; $(2)$ since the optical flow is the projection of the scene flow onto the image plane, the scene flow is reconstructed through triangulation using the optical flow information. Following Vedula~\etal's work and building upon developments in optical flow and stereo matching, other works proposed the use of variational methods to jointly estimate the scene flow and the 3D structure from stereo sequences~\cite{huguet07, pons07, wedel08, valgaerts10, wedel11, bashal13}. Other relevant works relied on rigidity assumptions of the scene structure and motion~\cite{vogel11, vogel15, vogel13, menze15}.

\noindent\textbf{Point-based scene flow.}~
Most non-learning scene flow methods are image-based. Nevertheless, we can interpret the seminal non-rigid iterative closest point (NICP) work by Amberg~\etal~\cite{NICP} and related non-rigid registration methods as scene flow estimators. NICP is an iterative method to deform a 3D template to fit scanned meshes. The algorithm uses a locally affine regularization based on the mesh topology to constrain the deformation field to be smooth. However, NICP relies on proper initialization, is sensitive to holes in the geometry, and is only suitable for small-scale differences between the template and the scanned mesh.

\subsection{Learning-based scene flow}
\noindent\textbf{Image-based scene flow.}~
Learning-based methods to estimate scene flow from monocular images have recently been proposed~\cite{Brickwedde2019, Yang_2020_CVPR, Hur_2020_CVPR}. Since the monocular scene flow problem is ill-posed by nature, most methods rely on 3D prior assumptions learned from data. Yang and Ramanan~\cite{Yang_2020_CVPR} proposed a learning method to lifting optical flow to scene flow from monocular images using optical expansion.

\noindent\textbf{Point-based scene flow.}~
There has been a great interest in estimating scene flow directly from point clouds. Recently, Dewan~\etal~\cite{dctb16} proposed to estimate rigid scene flow from point clouds. They proposed an energy minimization problem of a factor graph with hand-crafted signature of histograms of orientations (SHOT) descriptors~\cite{tombari10} for correspondence search. Later, Ushani~\etal~\cite{uwwe17} proposed a logistic classifier to predict if two columns of occupancy grids are in correspondence. Then, they formulated an expectation-maximization (EM) algorithm to estimate a locally rigid scene flow. Behl~\etal~\cite{bpdg19} proposed a method to jointly predict scene flow and 3D bounding boxes with their rigid body motions. Liu~\etal~\cite{liu19} proposed FlowNet3D to extract point features using PointNet++~\cite{qysg17} and to estimate the scene flow using a flow embedding layer. Wang~\etal~\cite{wang20} proposed FlowNet3D++ to improve on FlowNet3D by incorporating geometric constraints in the form of point-to-plane distance and angular alignment. Gu~\etal~\cite{gwwlw19} presented HPLFlowNet to predict the scene flow using bilateral convolutional layers (BCL) to project the point cloud onto a permutohedral lattice. Wu~\etal~\cite{wu19} proposed PointPWCNet, a coarse-to-fine approach to predict scene flow from point clouds. Mainly, they employed PointConv~\cite{wu19pointconv} to learn features from points efficiently. Mittal~\etal~\cite{mittal2019just} presented a method to fine-tune a pre-trained FlowNet3D model with real data using a self-supervised loss based on nearest neighbors and cycle consistency.

\subsection{Graph Laplacian-based methods}
The graph Laplacian is a basis for a variety of geometry processing tasks, and it has been widely used for mesh editing~\cite{Keenan2015, s05, bobenko2007, arap2007}, mesh registration~\cite{eisenberger2019}, physics-based modeling~\cite{keenan2019}, spectral graph analysis~\cite{Wang2018_eccv}, among others. It is generally applied to achieve a common goal: constrain the parameters of deformation to be smooth functions along a surface. However, the graph Laplacian has not yet been fully investigated for the point-based scene flow problem.

\subsection{Overview}
Current point-based scene flow approaches rely on deep networks to achieve state-of-the-art performance over classical methods. They are usually supervised from large-scale synthetic data and then fine-tuned on small-scale real-world datasets. Given the scarcity of large-scale, real-world data with scene flow annotations, recent works have shifted towards self-supervised methods. To regularize the scene flow, these methods perform iterative sampling and grouping of neighboring point features in different scales. Our approach differs from the previously mentioned works in that our objective function does not rely on recursive point feature sampling and grouping as in~\cite{wu19, mittal2019just}. Instead, we build an \textit{explicit} graph on the source point cloud to capture the topology and context information about how points are locally connected and how they should move through the graph Laplacian. We seek to explore traditional objectives that have been abandoned for supervised methods.

\section{Background}
In this section, we briefly summarize the graph Laplacian formation process. Let $\mathbf{G} {=} \{\mathbf{V}, \mathbf{E}\}$ be an undirected graph with a set of vertices $\mathbf{V}$ connected by a set of edges $\mathbf{E}$. For a given set of $\mathbf{V}$ and $\mathbf{E}$, the graph can be formally represented by its adjacency matrix $\mathbf{A} \in \mathbb{R}^{N \times N}$ which describes the vertex connectivity for $N$ vertices. The element $A_{ij}$ of $\mathbf{A}$ assume values $\{0, 1\}$. The value $A_{ij} {=} 0$ is assigned if the vertices $i$ and $j$ are not connected with an edge, and $A_{ij} {=} 1$, if these vertices are connected, that is

\startcompact{small}
\begin{equation}
\begin{aligned}
    A_{ij} \defeq
    \begin{cases}
        1,  & \text{if } (i, j) \in \mathbf{E}\\
        0,  & \text{if } (i, j) \notin \mathbf{E}.
    \end{cases}
\end{aligned}
\end{equation}
\stopcompact{small}

The weight matrix $\mathbf{W}$ is similar to the adjacency matrix $\mathbf{A}$ definition but it can convey extra contextual information, for example, the relative importance of the edge connections based on a distance metric. Another matrix that captures information from a graph is the degree matrix $\mathbf{D}$. For an undirected graph, $\mathbf{D}$ is a diagonal matrix whose elements $D_{ii}$ are equal to the sum of weights of all edges connected to the vertex $i$, that is, the sum of elements in its $i-th$ row

\startcompact{small}
\begin{equation}
\begin{aligned}
D_{ii} \defeq \sum_{j=1}^{N-1} W_{ij}.
\end{aligned}
\end{equation}
\stopcompact{small}

The graph Laplacian matrix $\mathbf{L}$ combines the weight matrix and the degree matrix as $\mathbf{L} \defeq \mathbf{D} - \mathbf{W}$.


\noindent The elements of $\mathbf{L}$ are nonnegative real numbers at the diagonal positions and nonpositive real numbers at the off-diagonal positions. For an undirected graph, $\mathbf{L}$ is symmetric, \ie $\mathbf{L} {=} \mathbf{L}^T$. For practical reasons, it is often advantageous to use a normalized $\mathbf{L}$~\cite{stankovic19}, defined as

\startcompact{small}
\begin{equation}
\begin{aligned}
\mathbf{L} \defeq \mathbf{D}^{-1/2} (\mathbf{D}-\mathbf{W}) \mathbf{D}^{-1/2}.
\end{aligned}
\end{equation}
\stopcompact{small}

\noindent The graph Laplacian $\mathbf{L}$ indicates how smooth a graph function is. It is the discrete version of the Laplacian for continuous spaces which is defined by the second derivative of a function. A smooth graph function does not drastically change in value from one vertex to another connected vertex. Thus, we seek to explore $\mathbf{L}$ as a regularization term to enforce an ``as-rigid-as-possible''~\cite{arap2007} scene flow.

\section{Method}
We introduce a method to estimate the scene flow given two consecutive point clouds from a dynamic scene. We constrain the scene flow with the graph Laplacian such that points that are close to each other move rigidly while the collection of all points move ``as-rigidly-as-possible''.

\subsection{Problem formulation}
Let $\mathbf{P}_{t-1} \in \mathbb{R}^{n_1 \times 3}$ be the 3D point cloud with $n_1$ points at time $t{-}1$ (source point cloud) and $\mathbf{P}_t \in \mathbb{R}^{n_2 \times 3}$ be the point cloud with $n_2$ points at time $t$ (target point cloud). To recover the scene flow, the source point cloud $\mathbf{P}_{t-1}$ should move close to the target point cloud $\mathbf{P}_{t}$. Therefore, a criterion for the fitting is that each point $\mathbf{p}_{t-1} {\in} \mathbf{P}_{t-1}$ should be as near as possible to its corresponding point in $\mathbf{P}_{t}$. In a rigid scene, a global rigid transformation is sufficient to recover the scene flow---all points $\mathbf{p}_{t-1}$ share the same transformation. However, most scenes we are interested in are non-rigid. For example, objects within a self-driving scenario are dynamic and have independent behaviors. A formulation to take into account all non-rigid motions is to solve for a translational vector $\mathbf{f} \in \mathbb{R}^{3}$ for each point $\mathbf{p}_{t-1} \in \mathbf{P}_{t-1}$. The collection of all translational vectors $\mathbf{f}$ is the scene flow $\mathbf{F} \in \mathbb{R}^{n_1 \times 3}$. Thus $\mathbf{P}_{t-1} {+} \mathbf{F}$ translates $\mathbf{P}_{t-1}$ towards $\mathbf{P}_{t}$. If the scene flow is projected onto an image plane the optical flow is granted for free.

\subsection{Data term}
An optimal scene flow estimation means an exact match between the source point cloud $\mathbf{P}_{t-1}$ and the target point cloud $\mathbf{P}_{t}$. However, real-world point clouds do not necessarily have the same number of points ($n_1$ is likely different from $n_2$) nor exact correspondences for exact matching. Thus, we define a data objective term $E_d$ to measure the quality of the point cloud matching as

\startcompact{small}
\begin{equation}
	\begin{aligned}
         E_d(\mathbf{F}) = \sum_{i=1}^{n_1} dist^2(\mathbf{p}_{{t-1}_i} + \mathbf{f}_i, \mathbf{P}_t),
	\end{aligned}
\label{eq:data_term}
\end{equation}
\stopcompact{small}

\noindent where the $dist(\cdot)$ function computes the distance to the closest corresponding point on $\mathbf{P}_{t}$. If all of the $\mathbf{f}_i$ are constrained to be the same, then minimizing the data term would solve for a global rigid transformation.
 
\subsection{Graph Laplacian term}
Each displacement vector $\mathbf{f}_i$ has three degrees of freedom in the optimization. If only using the data term $E_d$, the problem is underconstrained as there are as many translations as there are points in the source point cloud. To constrain the problem, we solve for a set of transformations that are ``as-rigid-as-possible''. We formulate the constraint as

\startcompact{small}
\begin{equation}
	\begin{aligned}
         E_\mathcal{L}(\mathbf{F}) = \sum_{\mathclap{\{i,j\} \in \mathbf{E}}} \|\mathbf{f}_i - \mathbf{f}_j\|_2^2,
	\end{aligned}
\label{eq:reg_term}
\end{equation}
\stopcompact{small}

\noindent where $\mathbf{E}$ is the set of edges of a graph $\mathbf{G}$ formed on $\mathbf{P}_{t-1}$.

There exist a variety of choices in which graphs are constructed from a set of points, including $k$-nearest neighbor ($k$-NN) graphs, $r$-neighborhood graphs, and ``self-tuning'' graphs~\cite{manor2004}. However, the two most commonly used are the $r$-neighborhood graph and the $k$-NN graph. In the $r$-neighborhood graph, the neighborhoods are restricted to lie inside a sphere with radius $r$. The $k$-NN graph does not define the neighborhoods with a fixed length-scale $r$ but rather by specifying for each point a set of $k$ nearest neighbors.

In this work, we focus on $k$-NN graphs as they are almost always preferred in practice over $r$-neighborhood graphs due to its simplicity and better sparsity and connectivity properties~\cite{calder2019}. Hence we construct a $k$-NN graph $\mathbf{G}$ on the source point cloud $\mathbf{P}_{t-1}$ with the purpose of leveraging its geometrical and topological information. Formally, given the set of $n_1$ points $\mathbf{P}_{t-1} {=} \{\mathbf{p}_{{t-1}_1}, \cdots, \mathbf{p}_{{t-1}_{n_1}}\}$, the undirected $k$-NN graph consists of the vertex set $\mathbf{V}$ and the edge set $\mathbf{E}$ which is a subset of $\mathbf{P}_{t-1} {\times} \mathbf{P}_{t-1}$. The vertices $\mathbf{p}_{{t-1}_i}$ and $\mathbf{p}_{{t-1}_j}$ are linked to form an edge if $\mathbf{p}_{{t-1}_i}$ is a $k$-nearest neighbor of $\mathbf{p}_{{t-1}_j}$ or vice versa.


Given $\mathbf{G}$, we compute the normalized graph Laplacian $\mathbf{L} \defeq \mathbf{D}^{-1/2} (\mathbf{D}-\mathbf{W}) \mathbf{D}^{-1/2}$, where $\mathbf{D}$ is the degree matrix and $\mathbf{W}$ is the weight matrix with elements $W_{ij}$ defined as

\startcompact{small}
\begin{equation}
\begin{aligned}
    W_{ij} = 
    \begin{cases}
        e^{-r^2_{ij}},    & \text{if } (i, j) \in \mathbf{E}, \\
        1,  & \text{if } i = j, \\
        0,  & \text{if } (i, j) \notin \mathbf{E},
    \end{cases}
\end{aligned}
\end{equation}
\stopcompact{small}

\noindent where $r$ is the Euclidean distance between the vertices $i$ and $j$ of $\mathbf{G}$. We use a weighted graph such that the edges convey information about the relative importance of their connection based on its distance. Finally, the term on Eq.~\ref{eq:reg_term} can be redefined in matrix form using the graph Laplacian  $\mathbf{L}$ as

\startcompact{small}
\begin{equation}
	\begin{aligned}
         E_\mathcal{L}(\mathbf{F}) = \tr(\mathbf{F}^T\mathbf{L}\mathbf{F}),
	\end{aligned}
\label{eq:reg_term_simp}
\end{equation}
\stopcompact{small}

\noindent where the symbol $\tr(\cdot)$ is the trace of a matrix. 

The full objective function is defined as

\startcompact{small}
\begin{equation}
	\begin{aligned}
         E(\mathbf{F}) = E_d + \alpha E_\mathcal{L},
	\end{aligned}
\label{eq:objective}
\end{equation}
\stopcompact{small}

\noindent where the first term minimizes the distance between the transformed source point cloud $\mathbf{P}_{t-1} {+} \mathbf{F}$ and the target $\mathbf{P}_t$. The idea is to search for the scene flow $\mathbf{F}$ that best transforms $\mathbf{P}_{t-1}$ towards $\mathbf{P}_{t}$. The second term is the graph Laplacian constraint meaning that the objective function should not change too much between nearby points, \ie ``as-rigid-as-possible'', and $\alpha$ is a weighting factor for the graph Laplacian regularizer. Note that the graph Laplacian $\mathbf{L}$ is constant throughout the minimization. 

\subsection{Point cloud correspondences}
The data term $E_d$ relies on a function $dist(\cdot)$ to compute the distance to the closest corresponding point on $\mathbf{P}_{t}$, \ie for every point on $\mathbf{P}_{t-1}$, the function has to find the closest point on $\mathbf{P}_{t}$. An inexpensive way to approximate the correspondences between point clouds is through the Chamfer distance function~\cite{fsg17}. It is the average matching distance to the nearest points, and it is defined as

\startcompact{small}
\begin{equation}
\begin{aligned}
    \mathcal{C}(\mathbf{P}_{t}, \mathbf{P}_{t-1}) \defeq
    \sum_{\mathclap{\mathbf{x}\in\mathbf{P}_{t}}} \min_{\mathbf{y}\in\mathbf{P}_{t-1}} \|\mathbf{x} - \mathbf{y}\|_2^2 + \sum_{\mathclap{\mathbf{y}\in\mathbf{P}_{t-1}}} \min_{\mathbf{x}\in\mathbf{P}_{t}} \|\mathbf{x} - \mathbf{y}\|_2^2.
\end{aligned}
\end{equation}
\stopcompact{small}

\noindent For each point, the Chamfer distance finds the nearest neighbor in the other point cloud and sums the squared distances. Since it is a function of point locations in $\mathbf{P}_{t}$ and $\mathbf{P}_{t-1}$, the Chamfer distance is a continuous and a piecewise smooth function. Hence differentiable almost everywhere and applicable as a loss function~\cite{fsg17}.

The proposed objective function can be optimized through gradient-descent algorithms. Thus it can be used \textit{with or without learning}. With learning, one can employ the proposed objective as a loss function for self-supervised learning. Without learning, one can recover scene flow directly from pairwise point clouds without any supervision.

\section{Experiments}
\subsection{Setup}
\label{subsection:setup}
\noindent\textbf{Datasets.}~ 
We used the following four datasets:

\noindent\textbf{1.}~\textbf{FlyingThings3D}~\cite{MIFDB16} is a large-scale synthetic dataset consisting of stereo and RGB-D images rendered from randomly moving 3D CAD objects. We used the preprocessed dataset released by~\cite{liu19}, where the RGB-D images were converted to point clouds and the optical flow to scene flow. It contains 19,967 train samples and 2,000 test samples.

\noindent\textbf{2.}~\textbf{KITTI Scene Flow}~\cite{mg15, mhg15} was designed to evaluate image-based scene flow methods on self-driving scenarios. Lidar point clouds were collected using the Velodyne HDL-64E sensor. They accumulated seven nearby point clouds and projected onto the images to densify the depth maps. We used the preprocessed dataset released by~\cite{liu19}, where they lifted the depth maps to points clouds and the optical flow to scene flow. It contains 100 train and 50 test samples.

\noindent\textbf{3.}~\textbf{Argoverse}~\cite{argoverse} is a new self-driving dataset, in the spirit of KITTI, but with more data and HD maps containing lane centerlines and ground height. However, scene flow annotations are not provided. To quantitatively evaluate our method, we created a dataset, ``\textit{Argoverse Scene Flow}'', based on the information provided in the Argoverse 3D Tracking v1.1 set. Specifically, we used the point clouds sensed from two Velodyne VLP-32 sensors, the vehicle poses and the 3D object tracks to lift pseudo scene flow annotations\footnote{\label{note1}Please refer to the supplementary material for details.}. It contains 2,691 train and 212 test samples.

\noindent\textbf{4.}~\textbf{nuScenes}~\cite{nuscenes2019} is a large-scale self-driving dataset featuring tracking annotations, map information, lidar point clouds collected with a Velodyne VLP-32 sensor, among others. However, as in the Argoverse dataset, scene flow annotations are not provided. We use the same Argoverse preprocessing steps to create the ``\textit{nuScenes Scene Flow}''\footnoteref{note1}. It contains 1,513 train samples and 310 test samples.

For a fair comparison with previous works, we also removed the ground points from the Argoverse Scene Flow and nuScenes Scene Flow datasets\footnoteref{note1}. Moreover, we noted that Argoverse and nuScenes might contain many rigid scenes where ICP-based methods would perform well on them. However, to avoid biasing to rigid scene flows, we filtered rigid scenes from Argoverse and nuScenes.

\begin{figure}[t]
	\centering
    \includegraphics[width=\linewidth,keepaspectratio]{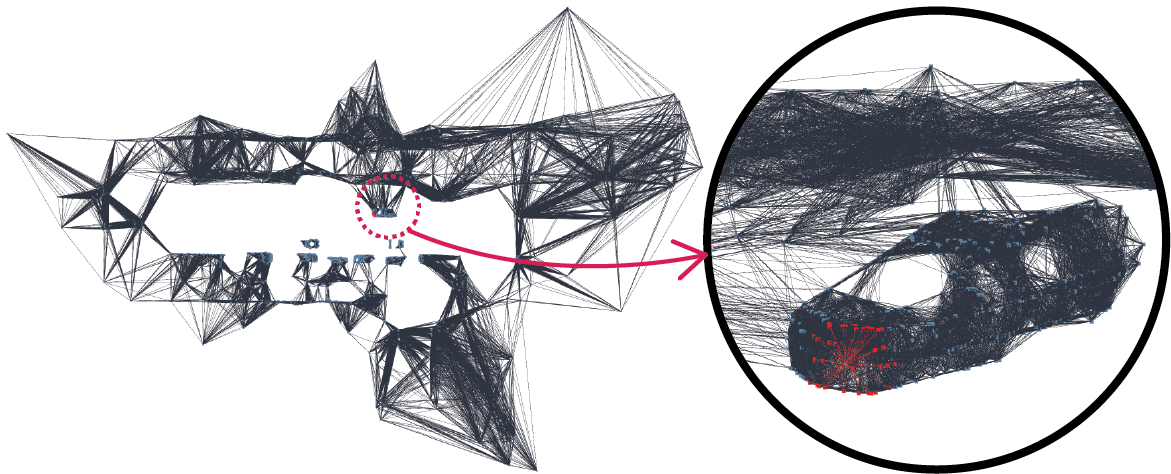}
\caption{\textbf{Implementation details.} Example of a $k$-NN graph formed on a point cloud from Argoverse Scene Flow. \textbf{Left:} Top view of the point cloud with its graph. \textbf{Right:} Zoomed-in area shows a vehicle and a local neighborhood region illustrated in red. The graph was built using fifty nearest neighbors (\ie $k{=}50$).}
\label{fig:graph}
\end{figure}

\begin{table*}[t!]
\begin{adjustbox}{width=\textwidth}
\begin{tabular}{@{}lrcccccccccccccccccc@{}}
\toprule
&\multicolumn{4}{c}{\thead{\textbf{FlyingThings3D}~\cite{MIFDB16} \\ \small \textit{\#Train: 19,967 $\vert$ \#Test: 2,000}}} &\hphantom &\multicolumn{4}{c}{\thead{\textbf{KITTI Scene Flow}~\cite{liu19} \\ \small \textit{\#Train: 100 $\vert$ \#Test: 50}}} &\hphantom &\multicolumn{4}{c}{\thead{\textbf{Argoverse Scene Flow}~\cite{argoverse} \\ \small \textit{\#Train: 2,691 $\vert$ \#Test: 212}}} &\hphantom &\multicolumn{4}{c}{\thead{\textbf{nuScenes Scene Flow}~\cite{nuscenes2019} \\ \small \textit{\#Train: 1,513 $\vert$ \#Test: 310}}} \\
\cmidrule{2-5} \cmidrule{7-10} \cmidrule{12-15} \cmidrule{17-20}
&\thead{$\mathcal{E}\downarrow$ \\ $(m)$} &\thead{$\%_5\uparrow$ \\ $(\%)$} &\thead{$\%_{10}\uparrow$ \\ $(\%)$} &\thead{$\theta_{\epsilon}\downarrow$ \\ $(rad)$} &&\thead{$\mathcal{E}\downarrow$ \\ $(m)$} &\thead{$\%_5\uparrow$ \\ $(\%)$} &\thead{$\%_{10}\uparrow$ \\ $(\%)$} &\thead{$\theta_{\epsilon}\downarrow$ \\ $(rad)$} &&\thead{$\mathcal{E}\downarrow$ \\ $(m)$} &\thead{$\%_5\uparrow$ \\ $(\%)$} &\thead{$\%_{10}\uparrow$ \\ $(\%)$} &\thead{$\theta_{\epsilon}\downarrow$ \\ $(rad)$} &&\thead{$\mathcal{E}\downarrow$ \\ $(m)$} &\thead{$\%_5\uparrow$ \\ $(\%)$} &\thead{$\%_{10}\uparrow$ \\ $(\%)$} &\thead{$\theta_{\epsilon}\downarrow$ \\ $(rad)$} \\ \toprule
\arrayrulecolor{sol_red}\toprule[0.5ex]
\tabitem\textbf{Deep Closest Point (DCP)}~\cite{Wang_2019_ICCV} &1.007 &0.17 &0.96 &1.213    &&0.638 &1.84 &3.91 &0.708  &&1.236 &0.09 &0.51 &1.453  &&1.277 &0.10 &1.04 &1.434 \\
\tabitem\textbf{FlowNet3D}~\cite{liu19} &0.134 &22.64 &54.17 &0.305     &&0.199 &10.44 &38.89 &0.386     &&0.455 &1.34 &6.12 &0.736       &&0.505 &2.12 &10.81 &0.620 \\
\tabitem\textbf{PointPWC-Net}~\cite{wu19} &\textbf{0.121} &\textbf{29.09} &\textbf{61.70} &\textbf{0.229}      &&\textbf{0.142} &\textbf{29.91} &\textbf{59.83} &\textbf{0.239}    &&\textbf{0.405} &\textbf{8.25} &\textbf{25.47} &\textbf{0.674}    &&\textbf{0.442} &\textbf{7.64} &\textbf{22.32} &\textbf{0.497} \\
\arrayrulecolor{sol_red}\toprule[0.5ex]
\arrayrulecolor{sol_green}\toprule[0.5ex]
\tabitem\textbf{Just Go with the Flow}~\cite{mittal2019just} &\multicolumn{4}{c}{---}      &&0.218 &10.17 &34.38 &0.254     &&0.542 &8.80 &20.28 &0.715        &&0.625 &6.09 &0.139 &0.432 \\
\tabitem\textbf{PointPWC-Net (self-sup. loss)}~\cite{wu19} &\multicolumn{4}{c}{---}     &&0.177 &13.29 &42.15 &0.272    &&0.409 &9.79 &\textbf{29.31} &0.643    &&0.431 &6.87 &22.42 &0.406 \\
\tabitem\textbf{Ours} 		        &\multicolumn{4}{c}{---}    &&\textbf{0.169} &\textbf{21.71} &\textbf{47.75} &\textbf{0.254}   &&\textbf{0.353} &\textbf{12.90} &{28.33} &\textbf{0.604} &&\textbf{0.284} &\textbf{14.50} &\textbf{35.46} &\textbf{0.363} \\
\arrayrulecolor{sol_green}\toprule[0.5ex]
\arrayrulecolor{sol_blue}\toprule[0.5ex]
\tabitem\textbf{Iterative Closest Point (ICP)}~\cite{ICP}            &0.412 &16.87 &34.56 &0.605     &&0.409 &5.24 &28.14 &0.608     &&0.438 &8.50 &24.70 &0.665     &&0.380 &15.03 &34.78 &0.450 \\
\tabitem\textbf{Non-rigid ICP (NICP)}~\cite{NICP}  &0.339 &14.05 &35.68 &0.480     &&0.338 &22.06 &43.03 &0.460    &&0.461 &4.27 &13.90 &0.741     &&0.402 &6.99 &21.01 &0.492 \\
\tabitem\textbf{PointPWC-Net (self-sup. loss)}~\cite{wu19} &0.433 &6.23 &19.46 &0.643     &&0.272 &16.98 &35.65 &0.338    &&0.466 &8.41 &22.62 &0.701     &&0.399 &8.31 &23.30 &0.454 \\
\tabitem\textbf{Ours} 		                &\textbf{0.259} &\textbf{16.30} &\textbf{41.60} &\textbf{0.369}     &&\textbf{0.093} &\textbf{64.76} &\textbf{82.13} &\textbf{0.137}   &&\textbf{0.257} &\textbf{25.26} &\textbf{47.50} &\textbf{0.467}    &&\textbf{0.288} &\textbf{20.19} &\textbf{43.59} &\textbf{0.337} \\
\arrayrulecolor{sol_blue}\bottomrule[0.5ex] 
\arrayrulecolor{black}\bottomrule
\end{tabular}
\end{adjustbox}
\vspace{0.01ex}
\caption{\textbf{Results.} Comparison with prior work on different datasets. Top section between \textcolor{sol_red}{\textbf{red bars}} shows \textit{off-the-shelf} \textbf{supervised learning methods}; the middle section between \textcolor{sol_green}{\textbf{green bars}} shows \textbf{self-supervised learning methods}; and the bottom section between \textcolor{sol_blue}{\textbf{blue bars}} shows \textbf{non-learning-based methods}. The off-the-shelf supervised learning methods were solely trained on synthetic datasets (models released by the authors). DCP was trained on ModelNet40; and FlowNet3D and PointPWC-Net were trained on FlyingThings3D. The self-supervised learning methods were trained on FlyingThings3D with supervision and then fine-tuned with self-supervision on each domain-specific dataset. $\mathcal{E}$ is the mean absolute distance error, or end-point error; $\%_5$ is the percentage of flow vectors where $\mathcal{E} < 0.05$ or $\delta < 5\%$, where $\delta$ is the percent error; $\%_{10}$ is the percentage of flow vectors where $\mathcal{E} < 0.1$ or $\delta < 10\%$; and $\theta_\epsilon$ is the mean angle error. $\downarrow$ means smaller values are better and $\uparrow$ means larger are better. All experiments were run with 2,048 points for each point cloud.}
\label{tab:main_table}
\end{table*}

\smallskip
\noindent\textbf{Metrics.}~ We used the following metrics: \textbf{1.} $\mathcal{E}$: the mean absolute distance error in meters, or end-point error; \textbf{2.} $\%_5$: the percentage of flow vectors where $\mathcal{E} {<} 0.05$ or $\delta {<} 5\%$; $\delta$ is the percent error; \textbf{3.} $\%_{10}$: the percentage of flow vectors where $\mathcal{E} {<} 0.1$ or $\delta {<} 10\%$; and \textbf{4.} $\theta_\epsilon$: the mean angle error in radians between the estimated and ground-truth scene flow.

\smallskip
\noindent\textbf{Implementation details.}~
We used the automatic differentiation in PyTorch~\cite{pytorch} to optimize our objective function using Adam~\cite{adam}. In the ``\textit{with learning}'' setting, we trained FlowNet3D on FlyingThings3D with supervision for 300 epochs, batch size of 16, and a learning rate $\gamma$ of $1\mathrm{e}{-3}$ and decaying it every 100 epochs by a factor of $0.1$. We further fine-tuned the model with the self-supervision of our proposed objective function on each real-world dataset. To avoid insufficient training, we used 1,000 epochs for fine-tuning, $\gamma$ of $1\mathrm{e}{-5}$, and the graph Laplacian regularizer weight $\alpha$ of $4.0$ (found by grid-search). In the ``\textit{without learning}'' setting, we ran the optimization for 1,500 epochs starting with a $\gamma$ of 0.1 and an $\alpha$ of $10.0$. We set the number of neighbors $k$ to 50 to construct the $k$-NN graph. We performed experiments on point clouds with 2,048, 4,096, and 8,192 points. Fig.~\ref{fig:graph} shows an example of a graph formed on a point cloud from Argoverse Scene Flow.

\smallskip
\noindent\textbf{Runtime.}~
It took \httilde{15} h to train FlowNet3D on FlyingThings3D for 300 epochs and \httilde{3} h to fine-tune it on KITTI Scene Flow with our self-supervised loss for 1,000 epochs. The fine-tuning on Argoverse Scene Flow took \httilde{3} days and on nuScenes took \httilde{2} days, both for 1,000 epochs. At test time, it took \httilde{36} ms for a single prediction using the learning-based approach. The non-learning version took \httilde{9} s to optimize a single scene flow. Experiments were run with 2,048 points, and on an NVIDIA Quadro P5000 GPU.

\subsection{Results}
Table~\ref{tab:main_table} compared our method to learning-based methods (\textit{with learning}): with supervision and with self-supervision, and against non-learning  methods (\textit{without learning}).

\subsubsection{With learning}
\noindent\textbf{With supervision.}~
We tested the generalization capabilities of three off-the-shelf state-of-the-art supervised methods: Deep Closest Point (DCP)~\cite{Wang_2019_ICCV}, FlowNet3D~\cite{liu19}, and PointPWC-Net~\cite{wu19}. DCP predicts a \textit{rigid} scene flow (\ie a global transformation). However, we used it as a baseline to verify that rigid approaches are not outperforming the non-rigid methods. We used the pre-trained DCP model on the ModelNet40 dataset and tested it on different datasets. Since DCP was trained on noise-free CAD objects, it could not generalize well to unseen data. ICP~\cite{ICP}, in the non-learning section of Table~\ref{tab:main_table}, outperformed it. 

FlowNet3D and PointPWC-Net predict the scene flow from pairwise point clouds. Both methods were trained with supervision on FlyingThings3D (\textit{without fine-tuning}) and tested on all datasets. PointPWC-Net outperformed FlowNet3D which demonstrates that it better generalizes to unseen data. However, its performance is far from optimum given that supervised methods still struggle when tested on datasets that are statistically different from the training set.

\smallskip
\noindent\textbf{With self-supervision.}~ 
We compared our method against two state-of-the-art self-supervised learning methods: Just Go with the Flow~\cite{mittal2019just} and PointPWC-Net~\cite{wu19}. To the best of our knowledge, these are the only approaches that tackled self-supervision for scene flow estimation from point clouds. In this work, we are interested in the performance of the loss function by itself. However, we trained FlowNet3D on FlyingThings3D to use it as a baseline due to its simplicity. Then, we performed self-supervised fine-tuning on the KITTI, Argoverse, and nuScenes Scene Flow datasets using the loss functions proposed by Mittal~\etal~\cite{mittal2019just}, Wu~\etal~\cite{wu19}, and ours. Our method outperformed the previous works showing that our proposed objective function can be successfully applied in self-supervised learning schemes.

\begin{table}[t!]
\begin{adjustbox}{width=\columnwidth}
\begin{tabular}{@{}lrcccccccc@{}}
\toprule
&\multicolumn{4}{c}{\textbf{Without the graph Laplacian }} &\hphantom &\multicolumn{4}{c}{\textbf{With the graph Laplacian}} \\
\cmidrule{2-5} \cmidrule{7-10}
&\thead{$\mathcal{E}\downarrow$ \\ $(m)$} &\thead{$\%_5\uparrow$ \\ $(\%)$} &\thead{$\%_{10}\uparrow$ \\ $(\%)$} &\thead{$\theta_{\epsilon}\downarrow$ \\ $(rad)$} &&\thead{$\mathcal{E}\downarrow$ \\ $(m)$} &\thead{$\%_5\uparrow$ \\ $(\%)$} &\thead{$\%_{10}\uparrow$ \\ $(\%)$} &\thead{$\theta_{\epsilon}\downarrow$ \\ $(rad)$} \\ \toprule
\tabitem\textbf{FlyingThings3D}              &0.636 &1.58 &5.83 &1.059  &&\textbf{0.259} &\textbf{16.30} &\textbf{41.60} &\textbf{0.369} \\
\tabitem\textbf{KITTI Scene Flow}              &0.882 &1.59 &4.12 &1.123  &&\textbf{0.093} &\textbf{64.76} &\textbf{82.13} &\textbf{0.137} \\
\tabitem\textbf{Argoverse Scene Flow}      &0.906 &1.49 &5.00 &1.179  &&\textbf{0.257} &\textbf{25.26} &\textbf{47.50} &\textbf{0.467} \\
\tabitem\textbf{nuScenes Scene Flow}   &1.080 &1.84 &5.11 &1.181  &&\textbf{0.288} &\textbf{20.19} &\textbf{43.59} &\textbf{0.337} \\
\bottomrule
\end{tabular}
\end{adjustbox}
\vspace{0.01ex}
\caption{\textbf{Influence of the graph Laplacian.} Effect of the graph Laplacian regularizer on different datasets. Experiments were run without learning and each point cloud with 2,048 points.}
\label{tab:ablation}
\vspace{-4mm}
\end{table}

\begin{figure}[t!]
	\centering
    \subfigure[Without the graph Laplacian]{
       	\;\; \includegraphics[width=100cm,height=3.1cm,keepaspectratio]{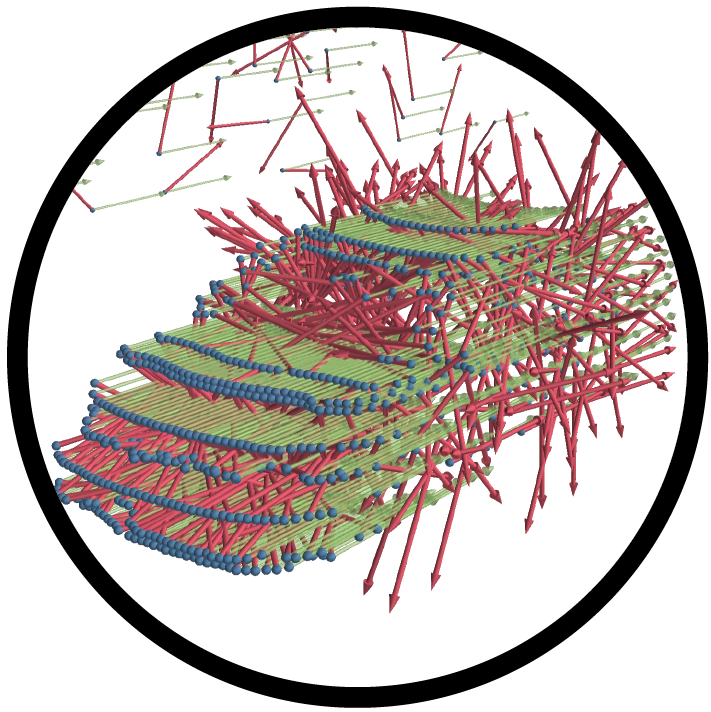} \;\;
      	\label{fig:without_laplacian}} \hspace{5mm}
	\subfigure[With the graph Laplacian]{
      	\includegraphics[width=100cm,height=3.1cm,keepaspectratio]{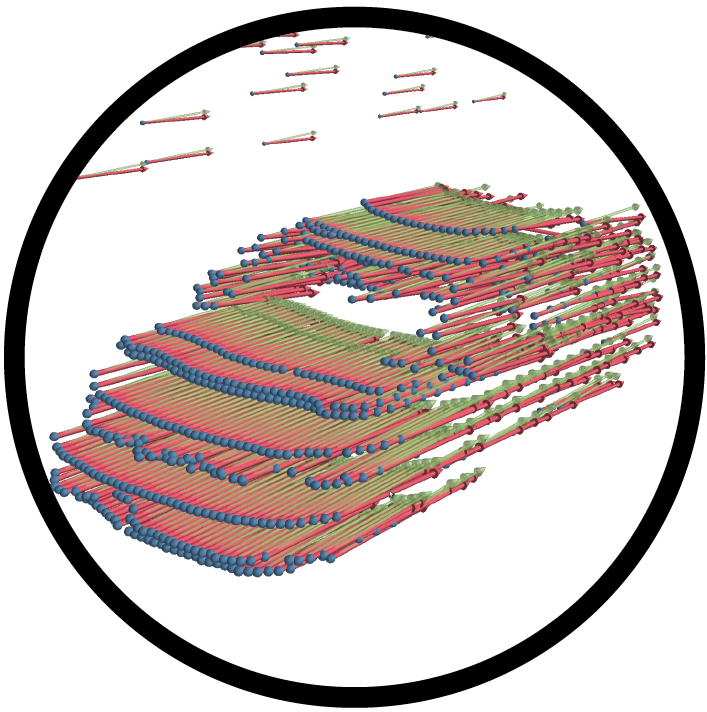}
    	\label{fig:with_laplacian}}
\caption{\textbf{Influence of the graph Laplacian.} Qualitative effect of the graph Laplacian regularizer on a scene from Argoverse Scene Flow. The zoomed in region shows a point cloud that was sampled from a vehicle (blue points). Green and red arrows are the ground-truth and the predicted scene flow by our non-learning method, respectively. Note how chaotic the scene flow is if not regularized.}
\label{fig:ablation}
\end{figure}

\begin{figure*}[t!]
	\centering
    \subfigure[Qualitative result on a scene from KITTI Scene Flow]{
       	\includegraphics[width=\linewidth,keepaspectratio]{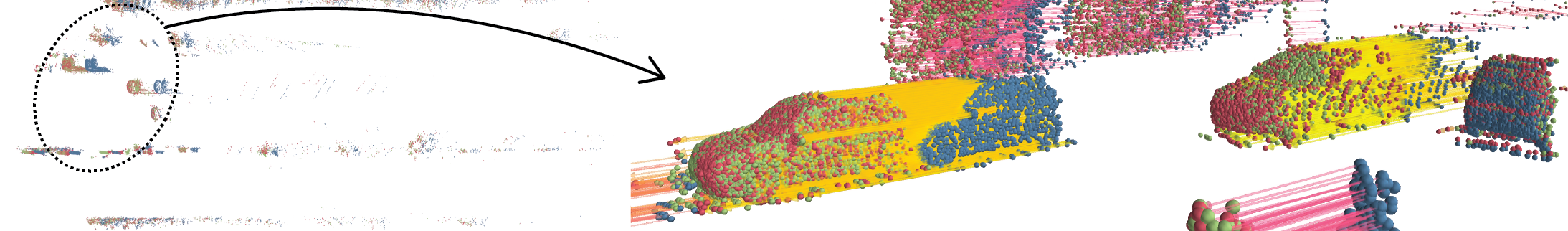}
      	\label{fig:kitti}}
	\subfigure[Qualitative result on a scene from Argoverse Scene Flow]{
      	\includegraphics[width=\linewidth,keepaspectratio]{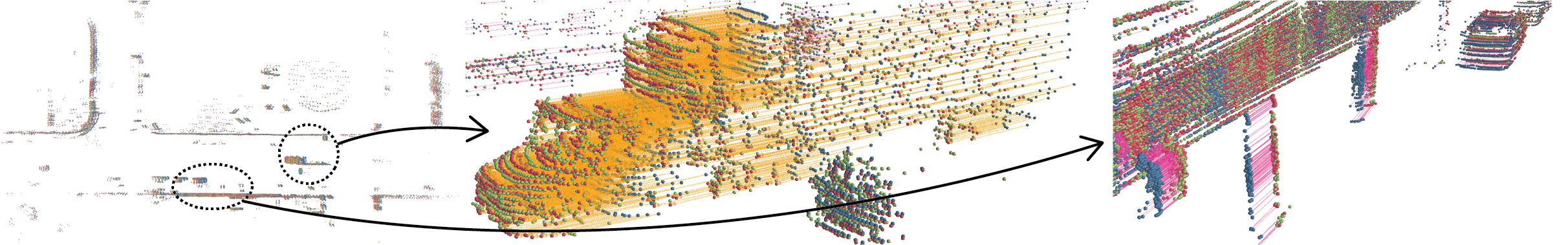}
    	\label{fig:argoverse}}
    \subfigure[Qualitative result on a scene from nuScenes Scene Flow]{
      	\includegraphics[width=\linewidth,keepaspectratio]{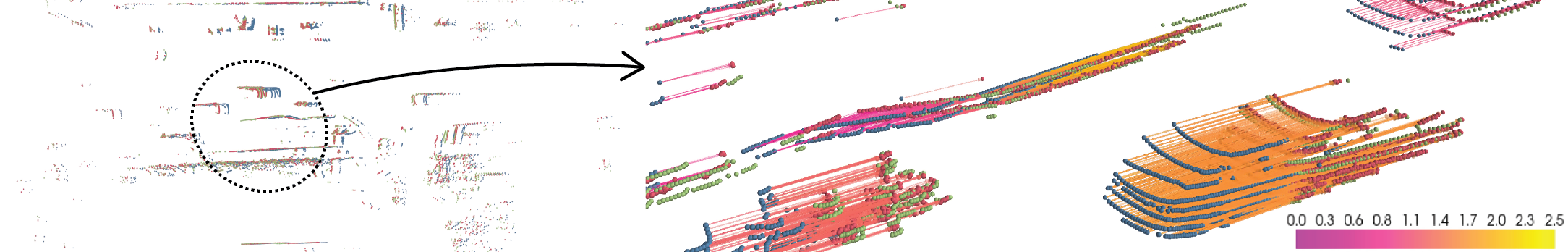}
    	\label{fig:nuscenes}}
\caption{\textbf{Qualitative results.} Results of our non-learning method on dynamic scenes from (a) KITTI, (b) Argoverse, and (c) nuScenes Scene Flow. \textbf{Left:} Top view of the whole scene. \textbf{Right:} Zoomed in areas. \textcolor{nordblue}{\textbf{Blue}} points are the source point cloud, $\mathbf{P}_{t-1}$, \textcolor{nordgreen}{\textbf{green}} points are the target point cloud, $\mathbf{P}_t$, \textcolor{nordred}{\textbf{red}} points are the translated source point cloud $\mathbf{P}_{t-1} {+} \mathbf{F}$. Arrows are the flows, $\mathbf{F}$, and its colors correspond to its magnitude (colorbar is shown in the bottom right). Note the different data modalities. For example, nuScenes has sparse measurements.}
\label{fig:qualitative}
\vspace{-4mm}
\end{figure*}

\subsubsection{Without learning}
We compared our non-learning method against three non-learning methods: ICP, NICP, and PointPWC-Net (only the self-supervised loss was employed). Results are shown in the blue section of Table~\ref{tab:main_table}. We did not compare against Mittal~\etal~\cite{mittal2019just} since their self-supervised loss is network dependent. Our method achieved better performance in all metrics. An advantage of using our non-learning method is that it only needs two point clouds without annotations to estimate scene flow robustly. Also, since the objective is optimized during runtime, extra priors might be integrated as regularizes. Fig.~\ref{fig:qualitative} shows visual results. Scene flow were recovered with high fidelity for challenging scenes\footnote{More examples in the supplementary material.}.


\begin{figure}[t!]
	\centering
    \includegraphics[width=\linewidth,keepaspectratio]{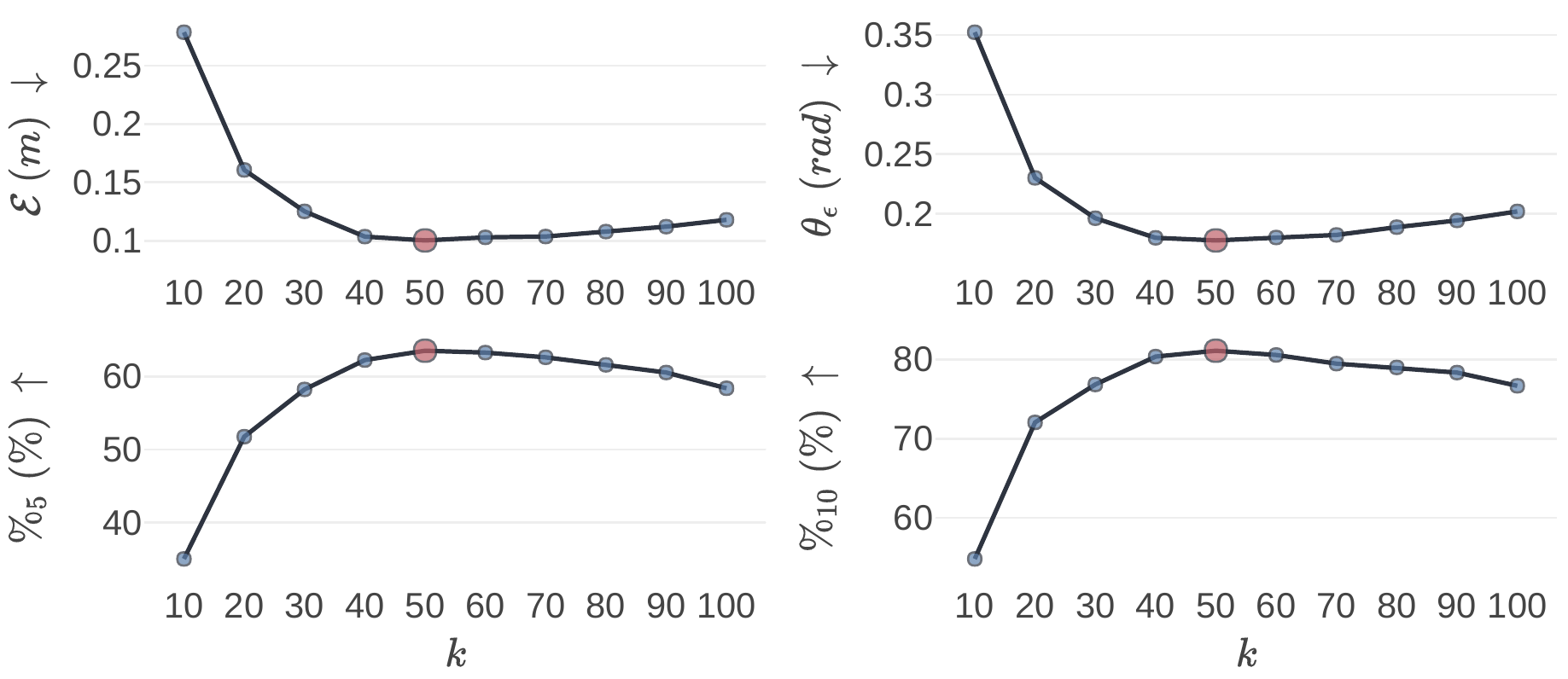}
\caption{\textbf{Influence of the graph Laplacian.} Performance of our approach on KITTI Scene Flow when varying the number of neighbors, $k$, to form the graph. Number of points was fixed to 2,048 for each point cloud. Red marks are the maxima or minima.}
\label{fig:graph_k}
\end{figure}

\begin{figure}[h]
	\centering
    \includegraphics[width=\linewidth,keepaspectratio]{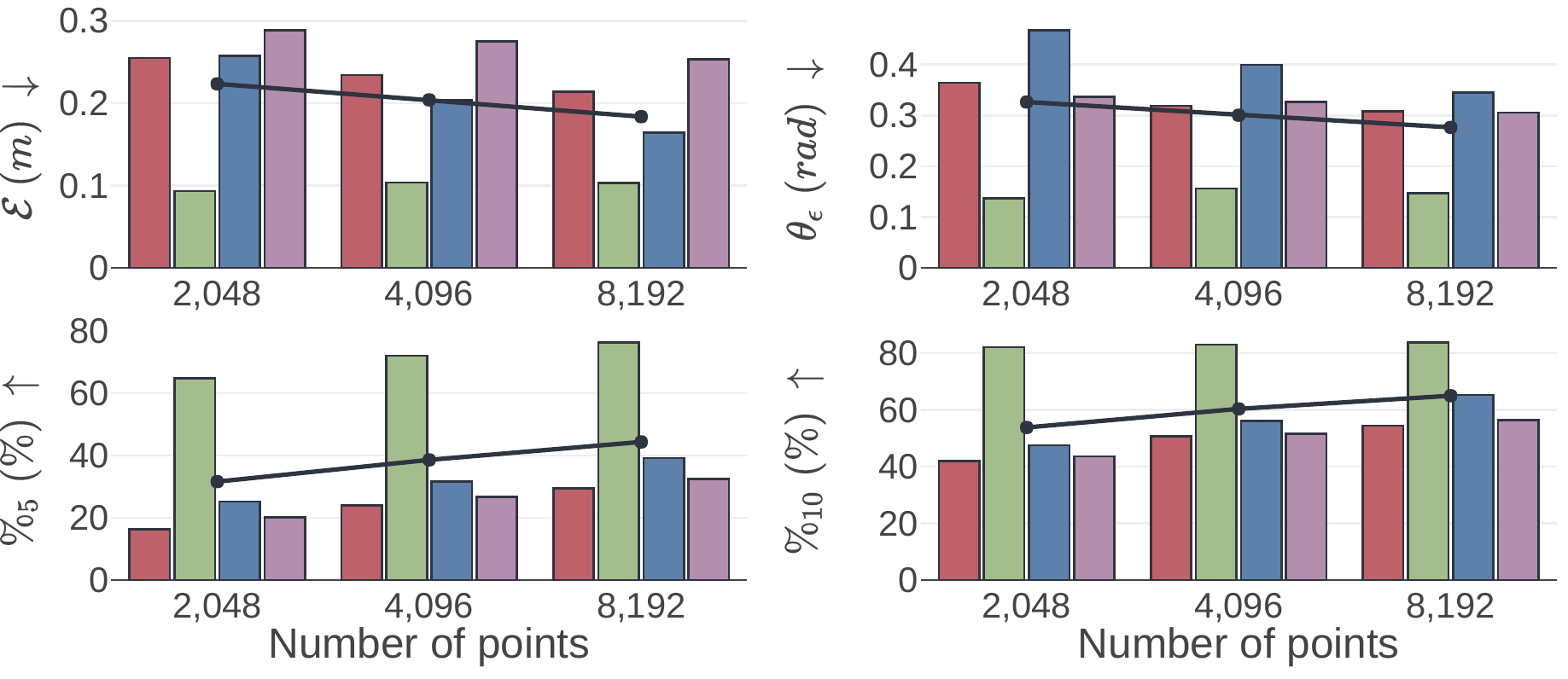}
\caption{\textbf{Influence of the number of points.} Performance of our method when varying the number of points. \legendsquare{fill=nordred} FlyingThings3D; \legendsquare{fill=nordgreen} KITTI Scene Flow; \legendsquare{fill=nordblue} Argoverse Scene Flow; \legendsquare{fill=nordpurple} nuScenes Scene Flow. Results improved with more points as shown in the trends.}
\label{fig:num_points}
\end{figure}

\subsubsection{Influence of the graph Laplacian}
\vspace{-2mm}
To understand the influence of the graph Laplacian regularizer, we removed it and evaluated the performance of the ablated model in Table~\ref{tab:ablation}. Visual results are shown in Fig.~\ref{fig:ablation}. The graph Laplacian regularizer drastically improved the performance in all metrics. We also tested the impact of the number of neighbors $k$ to form the graph. Fig.~\ref{fig:graph_k} shows the results when fixing the number of points to 2,048 and varying $k$. Experiments were performed on KITTI. The best performance was with $k{=}50$. Larger $k$'s did not increase the performance but slightly hurt it. This is explained by the fact that larger $k$'s encourages larger regions in the scene to move rigidly, \ie the scene flow will be less flexible.

\vspace{-4mm}
\subsubsection{Influence of the number of points}
We tested our non-learning model's performance when varying the number of points for both point clouds to 2,048, 4,096, and 8,192. In Fig.~\ref{fig:num_points}, we see that the performance grows as we increase the number of points. The denser the point clouds the easier it can be to search for better correspondences due to extra geometric information. However, a denser point cloud means a denser graph Laplacian due to more graph connectivity and a larger correspondence search space, thus increasing the computational complexity\footnote{Please refer to the supplementary material for extra runtime analysis.}. We also noted that the performance did not increase much for some metrics on the KITTI Scene Flow dataset when varying the number of points. This behavior suggests that the KITTI dataset preprocessed by~\cite{liu19} is close to performance saturation. Challenging datasets, such as FlyingThings3D, Argoverse, and nuScenes Scene Flow, are perhaps better suited to evaluate new scene flow methods.

\begin{figure}[t!]
	\centering
    \includegraphics[width=0.83\linewidth,keepaspectratio]{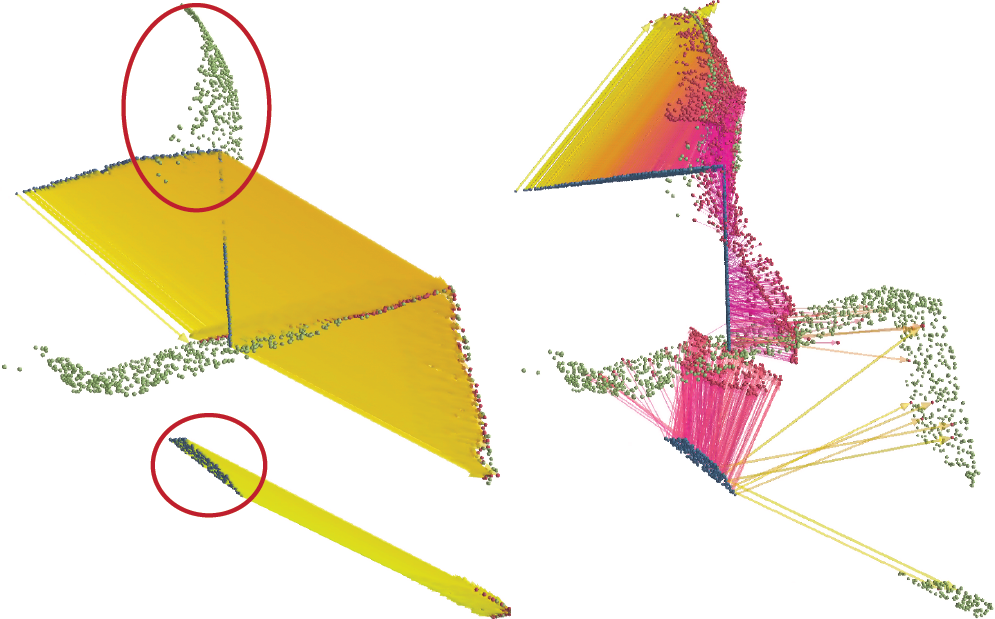}
\caption{\textbf{Limitations.} Failure case in a FlyingThings3D scene. \textcolor{nordblue}{\textbf{Blue}} is the source point cloud, \textcolor{nordgreen}{\textbf{green}} is the target point cloud, \textcolor{nordred}{\textbf{red}} is the translated source point cloud, and the arrows are the scene flow (colors correspond to its magnitude). \textbf{Left:} Shows the source, target, and ground-truth scene flow. \textbf{Right:} Shows the source, target, translated source, and the estimated scene flow by our method. The top red circle shows part of the target point cloud that is missing in the source point cloud. Thus the translated points got stuck in the wrong places due to bad correspondences. The bottom red circle shows part of the source point cloud that was wrongly translated due to the nearest-neighbors-based correspondence search.}
\label{fig:failure}
\end{figure}

\subsection{Limitations and Discussions}
We observed failure cases when the point clouds do not contain enough information to search for proper correspondences. This happens when the point clouds have large occlusions, holes, or missing parts, and these might lead to inaccurate scene flow estimation. Fig.~\ref{fig:failure} shows an example from FlyingThings3D where a change in visibility happened in the scene so that parts in the target point cloud is not visible in the source point cloud. Since we employed a nearest-neighbor-based method to search for correspondences, our approach will translate the source points to the nearest target points without knowledge of its local geometry and semantics. We also acknowledge that our proposed objective might not be optimal as there are several ways to construct the graph Laplacian. For example, one can use $r$-neighborhood graphs, ``self-tuning'' graphs, among others. However, we decided to formulate our method with the $k$-NN graph, which is simple and practical to use as a scene flow regularizer. Moreover, our non-learning method is orders of magnitude slower than our learning method (see runtime in~\ref{subsection:setup}). Nevertheless, it can be applied to cases that require robustness instead of optimal speed. If robustness is not an issue, but rather time, our proposed objective function can be used to train a self-supervised model and act as a surrogate of our non-learning method.

\begin{figure}[t!]
	\centering
    \includegraphics[width=\linewidth,keepaspectratio]{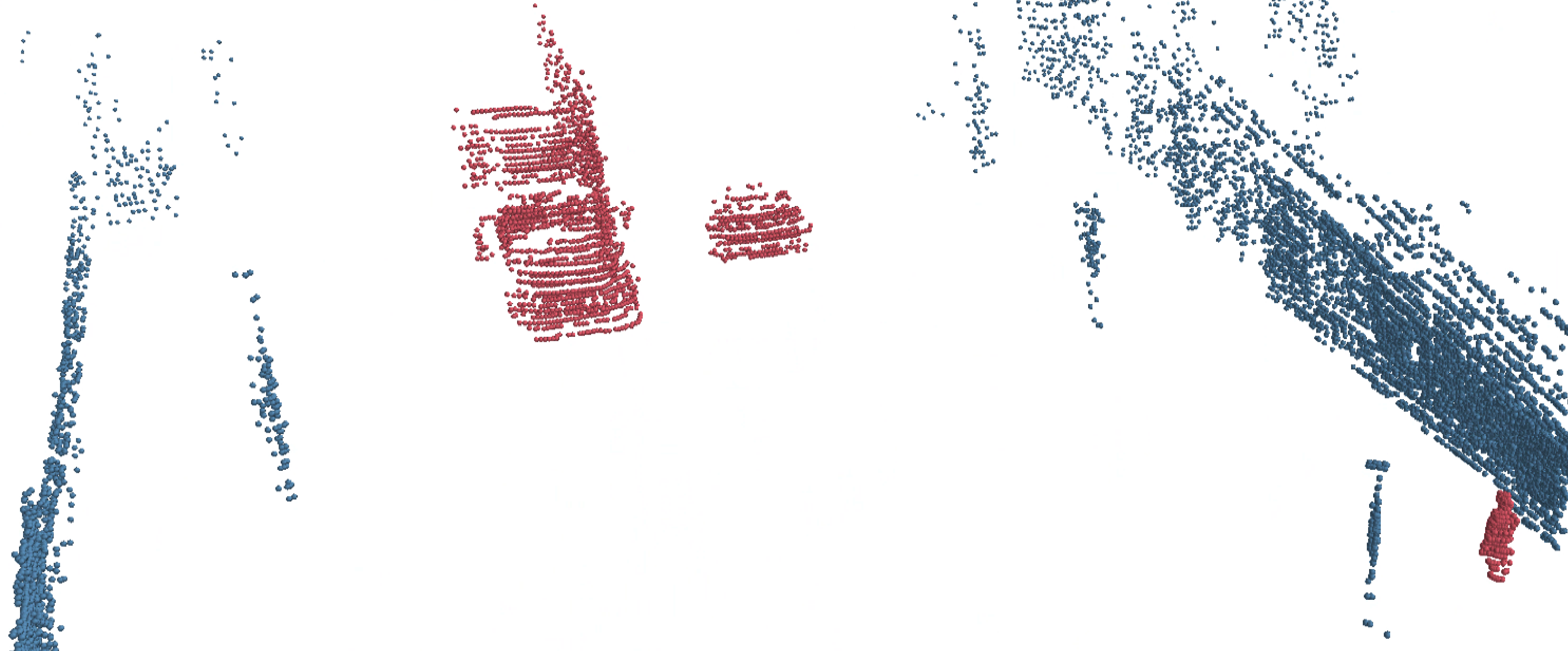}
\caption{\textbf{Motion segmentation.} The red points have larger motions relative to the blue points. A truck, car, and pedestrian (bottom right) were segmented from the Argoverse Scene Flow scene.}
\label{fig:segmentation}
\end{figure}

\subsection{Applications}
\noindent\textbf{Motion segmentation.}~ 
Discontinuities in the scene flow can help to segment point clouds into regions that correspond to different objects. Fig.~\ref{fig:segmentation} shows an example of a scene segmentation. We simply set a threshold to filter large motions in the estimated scene flow.

\smallskip
\noindent\textbf{Point cloud densification.}~
Fig.~\ref{fig:accumulation} shows that our method can be applied to densify point clouds from dynamic scenes. Point cloud densification might be helpful to create dense depth maps from lidar point clouds and images. Five adjacent frames from an Argoverse scene in each direction were used to densify the current frame. Each point cloud has about 65$\mathrm{e}$3 points. We visually compared our non-rigid densification against the original sparse point cloud and ICP. We did not use any semantic information nor a temporal consistency term across the frames for the registrations.

\section{Conclusion}
We presented a method to estimate the scene flow of dynamic scenes given a pair of point clouds. We proposed a simple and interpretable objective function to robustly approximate the scene flow with an ``as-rigid-as-possible'' regularizer based on the graph Laplacian. We successfully demonstrated that our objective function not only can be employed in self-supervised models when scene flow annotations are not available but also as a non-learning-based method in which the scene flow is optimized during runtime. Our approach outperformed the current self-supervised methods \textit{with or without learning}.

\blfootnote{\noindent\textbf{Acknowledgments.} We thank our Argo AI colleagues for their helpful suggestions. James Hays receives research funding from Argo AI, which is developing products related to the research described in this paper. In addition, the author serves as a principal scientist to Argo AI. The terms of this arrangement have been reviewed and approved by Georgia Tech in accordance with its conflict of interest policies.}

{\small
\bibliographystyle{ieee}
\bibliography{egbib}
}


\clearpage

\noindent\textbf{\large Supplementary Material}
\setcounter{section}{0}
\setcounter{figure}{0}

\section{Datasets} 
Here we provide additional information to supplement our main submission regarding the dataset preprocessing described in Section 5.1.

\subsection{Argoverse Scene Flow}
Scene flow annotations are not provided in the Argoverse~\cite{argoverse} dataset. To quantitatively evaluate our method, we created a dataset, ``\textit{Argoverse Scene Flow}'', based on the information provided in the Argoverse 3D Tracking v1.1 set. We used lidar point clouds sensed from two Velodyne VLP-32 sensors, six degrees of freedom (6DoF) vehicle poses, and the 3D object tracks to lift pseudo scene flow annotations. The 3D object tracks consist of bounding cuboids and poses for all objects of interest---both dynamic and static.

Given two consecutive lidar point clouds, we separated the rigid and non-rigid objects for both of them using the object track information. Then, we registered the rigid parts of both point clouds using the 6DoF vehicle pose, and the non-rigid segments were registered using the object poses. Thus the translational vectors can be recovered from the relative transformations to create the pseudo scene flow. Moreover, we used the ground height map information available in the Argoverse dataset to perform ground-points removal.


\subsection{nuScenes Scene Flow}
Scene flow annotations are also not provided in the nuScenes~\cite{nuscenes2019} dataset. We used the same Argoverse preprocessing steps to create the ``\textit{nuScenes Scene Flow}''. However, since nuScenes does not provide ground height maps, we used RANSAC to remove the ground points\footnote{We used the Open3D library~\cite{open3d} to perform ground-point removal.}.

\section{Qualitative Results} 
\begin{figure*}[t!]
	\centering
    \subfigure[Without learning: Qualitative result on a scene from KITTI Scene Flow]{
       	\includegraphics[width=\linewidth,keepaspectratio]{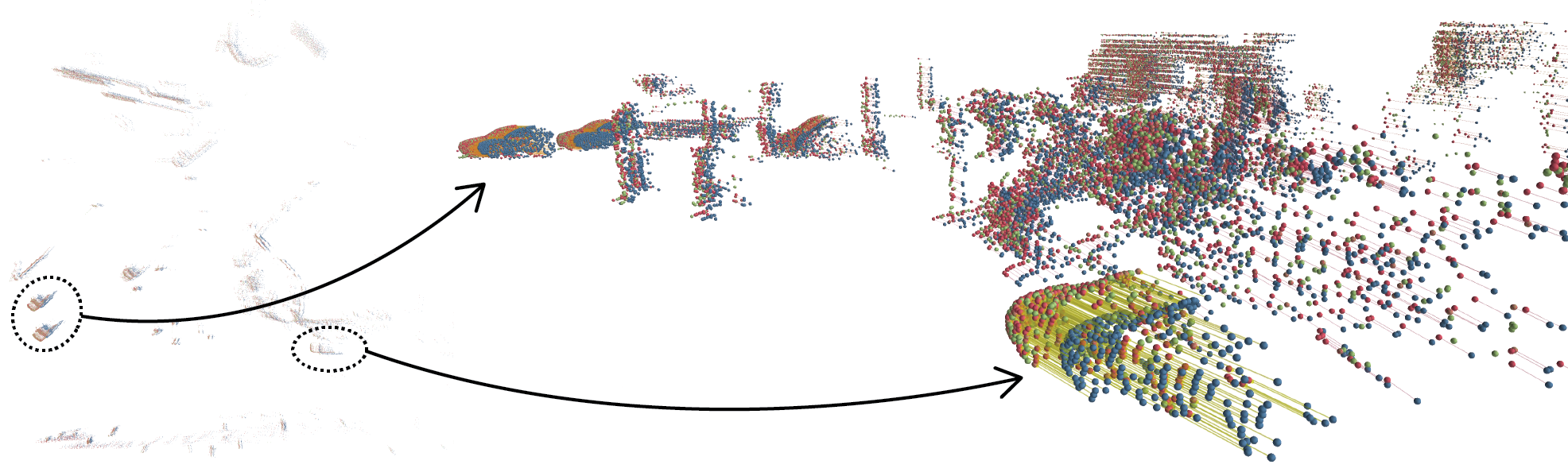}
      	\label{fig:kitti_no_learn}}
	\subfigure[With learning: Qualitative result on a scene from KITTI Scene Flow]{
      	\includegraphics[width=\linewidth,keepaspectratio]{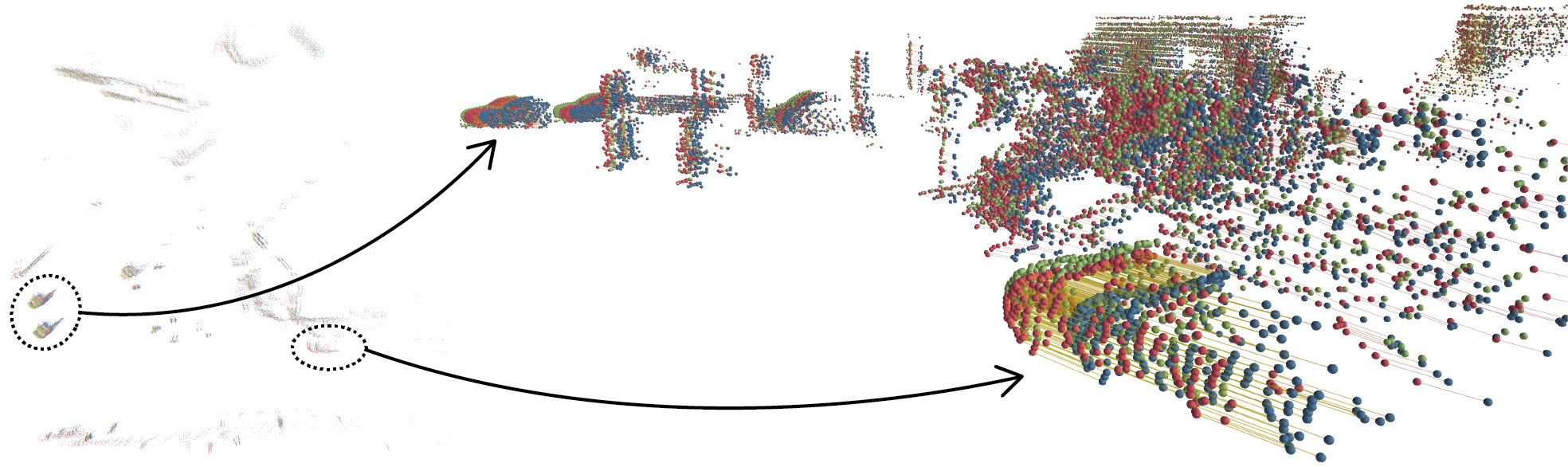}
    	\label{fig:kitti_self_learn}}
\caption{\textbf{Qualitative results.} Results of our scene flow method on a dynamic scene from KITTI Scene Flow (a) without learning and (b) with self-supervised learning. \textbf{Left:} Top view of the whole scene. \textbf{Right:} Zoomed in areas. \textcolor{nordblue}{\textbf{Blue}} points are the source point cloud, $\mathbf{P}_{t-1}$, \textcolor{nordgreen}{\textbf{green}} points are the target point cloud, $\mathbf{P}_t$, \textcolor{nordred}{\textbf{red}} points are the translated source point cloud $\mathbf{P}_{t-1} {+} \mathbf{F}$. Arrows are the flows, $\mathbf{F}$, and its colors correspond to its magnitude (yellowish is for bigger magnitudes). Note the better registration of our non-learning method.}
\label{fig:qualitative_kitti}
\vspace{-4mm}
\end{figure*}

\begin{figure*}[t!]
	\centering
    \subfigure[Without learning: Qualitative result on a scene from Argoverse Scene Flow]{
      	\includegraphics[width=\linewidth,keepaspectratio]{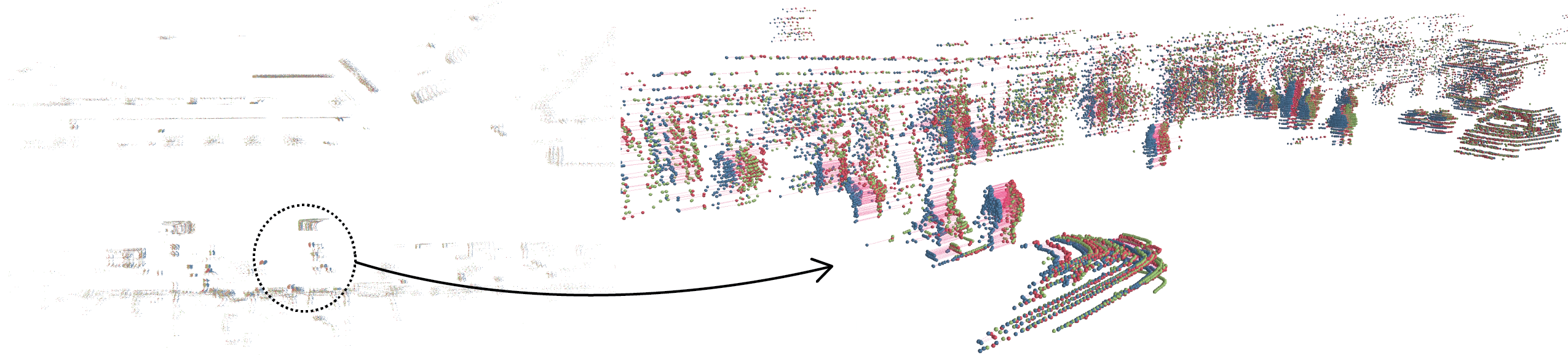}
    	\label{fig:argoverse_no_learn}}
    \subfigure[With learning: Qualitative result on a scene from Argoverse Scene Flow]{
      	\includegraphics[width=\linewidth,keepaspectratio]{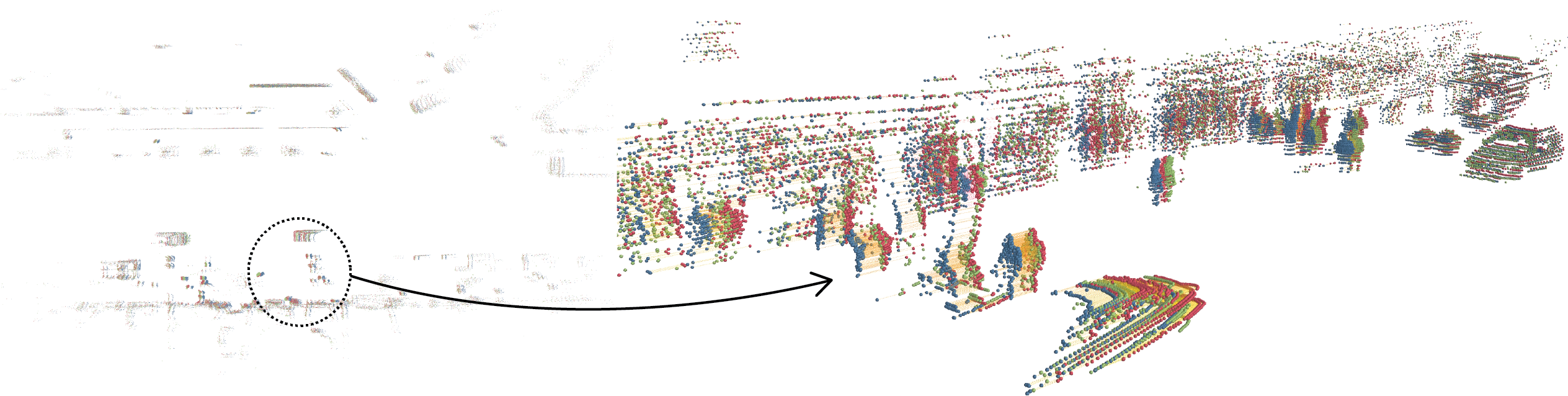}
    	\label{fig:argoverse_self_learn}}
\caption{\textbf{Qualitative results.} Results of our scene flow method on a dynamic scene from Argoverse Scene Flow (a) without learning and (b) with self-supervised learning. \textbf{Left:} Top view of the whole scene. \textbf{Right:} Zoomed in areas. \textcolor{nordblue}{\textbf{Blue}} points are the source point cloud, $\mathbf{P}_{t-1}$, \textcolor{nordgreen}{\textbf{green}} points are the target point cloud, $\mathbf{P}_t$, \textcolor{nordred}{\textbf{red}} points are the translated source point cloud $\mathbf{P}_{t-1} {+} \mathbf{F}$. Arrows are the flows, $\mathbf{F}$, and its colors correspond to its magnitude (yellowish is for bigger magnitudes). Note the better registration of our non-learning method.}
\label{fig:qualitative_argoverse}
\vspace{-4mm}
\end{figure*}

\begin{figure*}[t!]
	\centering
    \subfigure[Without learning: Qualitative result on a scene from nuScenes Scene Flow]{
      	\includegraphics[width=\linewidth,keepaspectratio]{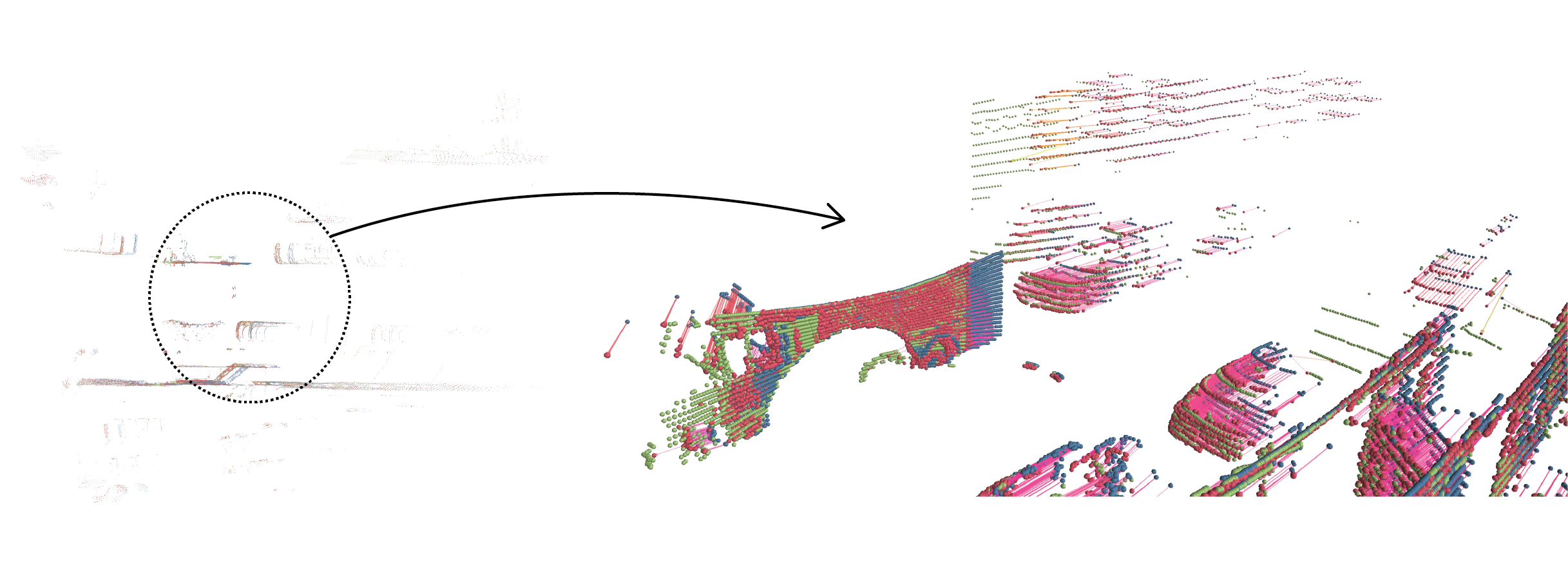}
    	\label{fig:nuscenes_no_learn}}
    \subfigure[With learning: Qualitative result on a scene from nuScenes Scene Flow]{
      	\includegraphics[width=\linewidth,keepaspectratio]{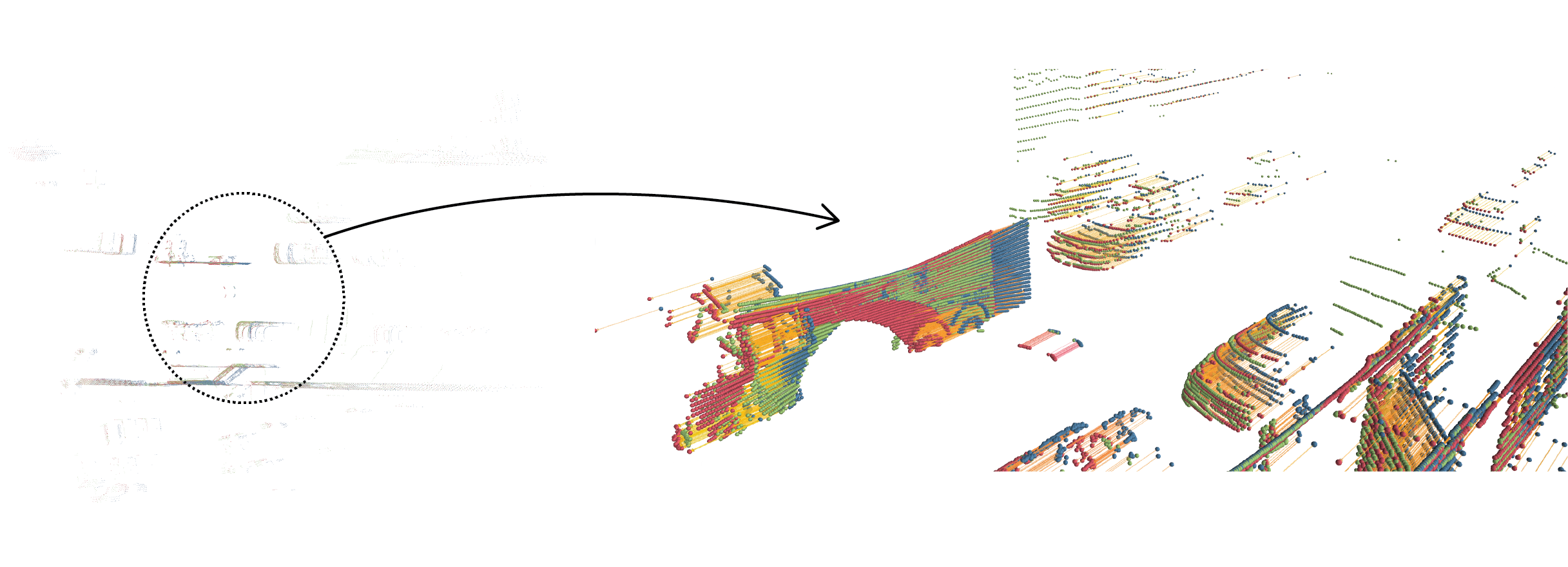}
    	\label{fig:nuscenes_self_learn}}
\caption{\textbf{Qualitative results.} Results of our scene flow method on a dynamic scene from nuScenes Scene Flow (a) without learning and (b) with self-supervised learning. \textbf{Left:} Top view of the whole scene. \textbf{Right:} Zoomed in areas. \textcolor{nordblue}{\textbf{Blue}} points are the source point cloud, $\mathbf{P}_{t-1}$, \textcolor{nordgreen}{\textbf{green}} points are the target point cloud, $\mathbf{P}_t$, \textcolor{nordred}{\textbf{red}} points are the translated source point cloud $\mathbf{P}_{t-1} {+} \mathbf{F}$. Arrows are the flows, $\mathbf{F}$, and its colors correspond to its magnitude (yellowish is for bigger magnitudes). Note the better registration of our non-learning method, especially in the arch structure in the middle where a wrong scene flow was predicted by our self-supervised method.}
\label{fig:qualitative_nuscenes}
\vspace{-4mm}
\end{figure*}

Here we provide additional qualitative results to Section 5.2 to supplement our main submission.
Fig.~\ref{fig:qualitative_kitti}--\ref{fig:qualitative_nuscenes} show visual results for our non-learning and self-supervised learning methods evaluated on the KITTI, Argoverse, and nuScenes Scene Flow datasets. The results show that our non-learning method provides scene flow estimation with higher fidelity than the self-supervised method.

\begin{figure}[t!]
	\centering
    \includegraphics[width=0.9\linewidth,keepaspectratio]{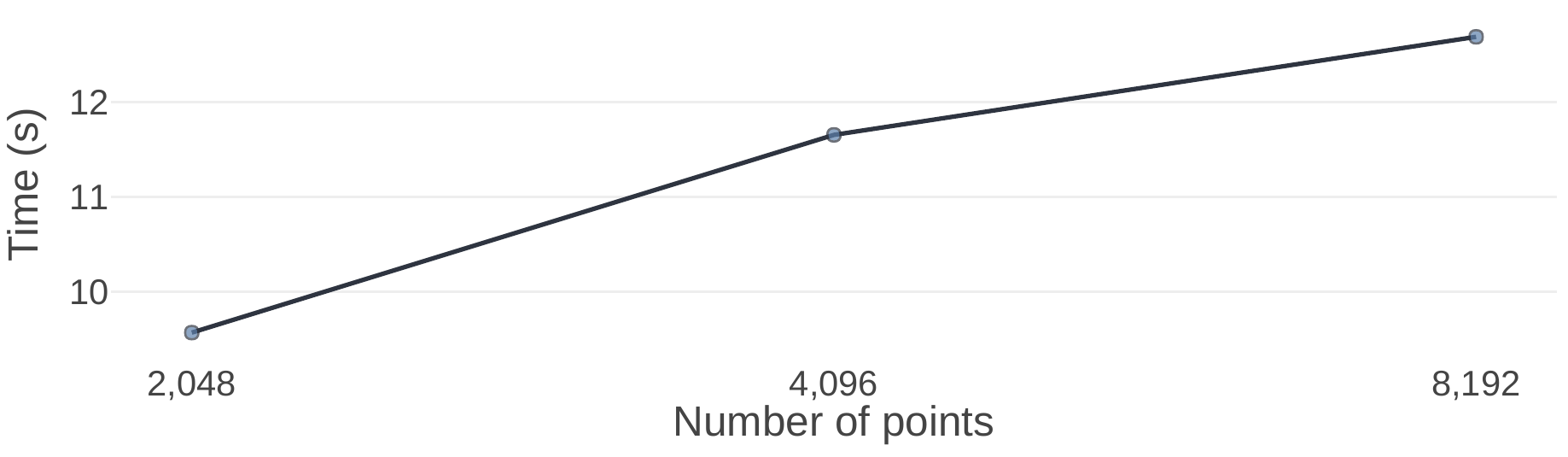}
\caption{\textbf{Influence of the number of points.} Approximated runtime of our non-learning method. Time corresponds to 1,500 iterations that is needed for converging.}
\label{fig:time}
\end{figure}

\section{Runtime}
Here we provide new results to Section 5.2.4 to supplement our main submission. We tested our non-learning model's performance when varying the number of points for both point clouds to 2,048, 4,096, and 8,192. Fig.~\ref{fig:time} shows that the time grows as we increase the number of points. For this evaluation, we assumed
that both point clouds, source and target, have the same number of points. We tested the runtime on a single NVIDIA Quadro P5000 GPU.

\end{document}